\documentclass[runningheads]{llncs}
\usepackage{comment}
\usepackage{graphicx}
\usepackage{multirow} 
\usepackage{booktabs} 
\usepackage{float} 
\newcommand*\samethanks[1][\value{footnote}]{\footnotemark[#1]}

\begin{document}
\title{Resolution-Agnostic Neural Compression for High-Fidelity Portrait Video Conferencing via Implicit Radiance Fields
% \thanks{Supported by organization x.}
}

\def\CICAISubNumber{50}  % Insert your submission number here
%\begin{comment}
% \titlerunning{IFTC2023 submission ID \CICAISubNumber} 
% \authorrunning{IFTC2023 submission ID \CICAISubNumber} 
% \author{Anonymous IFTC submission}
% \institute{Paper ID \CICAISubNumber}
%\end{comment}
%******************

%\begin{comment}
\titlerunning{NeRF-based Video Compression}
% \author{Yifei Li\inst{1}\orcidID{0000-1111-2222-3333} \and
% Xiaohong Liu\inst{1}\orcidID{1111-2222-3333-4444} \and
% Yicong Peng\inst{1}\orcidID{2222--3333-4444-5555} \and
% Guangtao Zhai\inst{1} \and
% Jun Zhou\inst{1}
% }
\author{Yifei Li\inst{1} \and
Xiaohong Liu\inst{1}\thanks{Corresponding author} \and
Yicong Peng\inst{1} \and
Guangtao Zhai\inst{1} \and
Jun Zhou\inst{1}\samethanks
}
\authorrunning{Y. Li et al.}
% Princeton NJ 08544 下面地址邮编的原始例子
% \institute{Shanghai Jiao Tong University, Shanghai, China \and
\institute{$^1$Shanghai Jiao Tong University, Shanghai, China}
%\end{comment}
%******************
\maketitle              % typeset the header of the contribution

\begin{figure}
    \includegraphics[width=\textwidth]{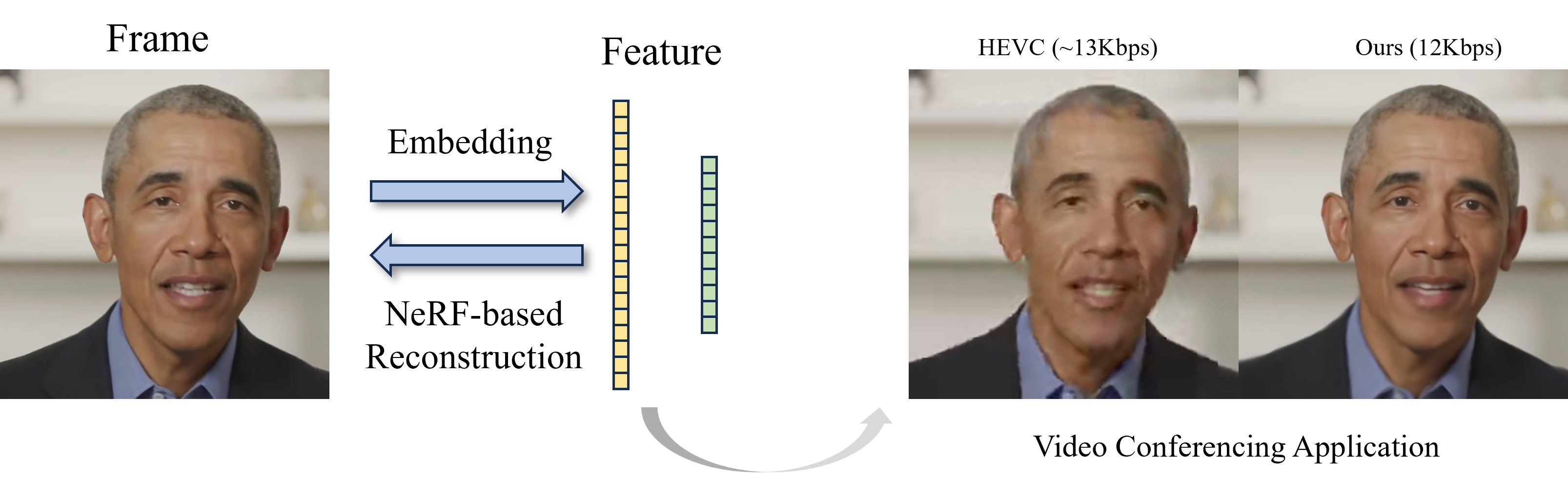}
    \caption{Illustration of our NeRF-based video compression. The core idea of our framework is frame-feature substitution for extremely low bandwidth. With NeRF-based face reconstruction model ensuring high-fidelity portrait generation, our framework shows significant compression performance for video conferencing application.} 
    \label{overview}
\end{figure}

\begin{abstract}
Video conferencing has caught much more attention recently. High fidelity and low bandwidth are two major objectives of video compression for video conferencing applications. Most pioneering methods rely on classic video compression codec without high-level feature embedding and thus can not reach the extremely low bandwidth. Recent works instead employ model-based neural compression to acquire ultra-low bitrates using sparse representations of each frame such as facial landmark information, while these approaches can not maintain high fidelity due to 2D image-based warping. In this paper, we propose a novel low bandwidth neural compression approach for high-fidelity portrait video conferencing using implicit radiance fields to achieve both major objectives. We leverage dynamic neural radiance fields to reconstruct high-fidelity talking head with expression features, which are represented as frame substitution for transmission. The overall system employs deep model to encode expression features at the sender and reconstruct portrait at the receiver with volume rendering as decoder for ultra-low bandwidth. In particular, with the characteristic of neural radiance fields based model, our compression approach is resolution-agnostic, which means that the low bandwidth achieved by our approach is independent of video resolution, while maintaining fidelity for higher resolution reconstruction. Experimental results demonstrate that our novel framework can (1) construct ultra-low bandwidth video conferencing, (2) maintain high fidelity portrait and (3) have better performance on high-resolution video compression than previous works.

\end{abstract}

\section{Introduction}
Video conferencing enables individuals or groups to participate in a virtual meeting by using video, which has caught much more attention since the online lifestyle becomes prevalent. Nowadays, the demand for video conferencing with the large amount of simultaneous users also determines its extremely low bandwidth limitations in application, which relies more heavily on efficient video compression technologies. Video compression aims to reduce video bandwidth while maintaining high fidelity. Over the past several decades, the dominant video compression methods are based on classic video compression frameworks, such as H.262, AVS \cite{avs}, H.264, HEVC \cite{hevc}, and VVC \cite{vvc}, which have achieved significant results. However, most classic methods reducing redundancy fully based on images and pixels without high level feature coding, thus can not reach the extremely limited low bandwidth while maintaining acceptable results in present video conferencing scenarios. 

% Recently, with the development of deep learning and neural network, several neural approaches have been proposed for video compression, 
With the development of computer science and deep learning, video, as the main medium for simultaneous auditory and visual outputs, has received extensive attention for its applications and research in related fields \cite{ftvos,3dpf,vigt,votm,aimfps,m3dsav,refam,cavs,srr}. At the same time, with the development of computer vision and graphics \cite{slpg,adcsu,oivc,mdas}, there are more and more neural network based methods targeting the resolution and quality of videos \cite{vfit,luvs,vfi,vsmmi,etvs,rms,TransMRSR,ovss,balof,sofnet}. Among them, the field of neural video compression has attracted much attention, where some neural compression methods \cite{fomm,bilayer,osfv,Maxime_low_bandwidth_video_chat,neural_compression_multi_views} leverage face image generative models to deliver extreme compression by reconstructing video frame from a high-level feature, such as motion keypoints \cite{osfv,Maxime_low_bandwidth_video_chat,ultra_low_fomm}. 
% Given a video clip, rather than simply based on pixels, these methods first extract corresponding feature embedding for each frame using self-supervised or pre-trained deep model as encoder. Then the high-level features substitute frames for transmission, which is the key to ultra-low bandwidth. At the receiver, portrait frames are reconstructed from features using generative models as decoder.
Specifically, most previous works use 2D warping based synthesis models to reconstruct portrait images. These warping methods deliver good reconstructions when the difference between the reference and target images is small, but they fail (possibly catastrophically) when there is large head pose movement or occlusion. 
% In this case, they have poor performance of both low-frequency content (e.g., the presence of a hand is completely omitted from a frame) and high-frequency content (e.g., details of clothing and facial hair). 
As a result, lacking of 3D representation, these warping based compression frameworks are not robust in maintaining high fidelity for some cases. Furthermore, most of these generative approaches have restrictions on input resolution (e.g., usually $256\times 256$), which means when it comes to high resolution applications (e.g., typical video conferencing are designed for HD videos), corresponding neural compression will not work. Meanwhile, with the superior capability in multi-view image synthesis of Neural Radiance Fields (NeRF) \cite{nerf}, several feature-conditioned dynamic neural radiance fields \cite{adnerf,4d_facenerf,dfa-nerf,neural_volume} have be proposed for talking head and dynamic face reconstruction. Rather than 2D warping, these models propose to use neural radiance fields to reconstruct portrait scene and represent the dynamics (e.g., expressions and head motion) as high-level features. Thanks to implicit 3D representation and volume rendering, these works are capable of producing natural portraits with high fidelity and more specifics (e.g., illumination and reflection) even in large movements. Nevertheless, to the best of our knowledge, the applications of such NeRF-based reconstruction model have not been delivered to neural video compression or video conferencing.

To address the defects of classic video codec and previous neural model-based compression and preserve both high fidelity and ultra-low bandwidth, we propose to leverage Neural Radiance Fields (NeRF) \cite{nerf} to reconstruct portrait in implicit 3D space for model-based neural compression and video conferencing. Specifically, we propose a novel neural compression framework using implicit neural radiance fields. At the sender, instead of using warping keypoints, we leverage 3D Morphable Face Models (3DMMs) \cite{3dmm} to extract facial expression feature and head pose from portrait frame. Due to its disentanglement of face attributes as a 3D representation, 3DMMs can gain control of face synthesis better. Besides, to obtain higher-level information representation and better compression performance, we propose to employ an attention-based model \cite{attention} as encoder for feature embedding, which is called \textit{fine-tuning embedding}. Before the features substituting frames to be transmitted, entropy coding as a lossless coding strategy is employed to compress the features further. Once the features have been received at the receiver, we leverage the feature-conditioned dynamic neural radiance fields to reconstruct the portrait video. It's worth noting that we refer to \cite{adnerf}, which has desirable performance in both face and torso rendering, and replace the audio feature with expression feature to build the face reconstruction model employed in our approach. We conduct extensive experiments in both quantitative and qualitative aspects with comparisons to classic video codec and previous model-based video compression. We demonstrate that our approach can reach extremely low bandwidth while maintaining high fidelity for video conferencing application. Furthermore, thanks to the characteristic of NeRF rendering with unlimited resolution \cite{nerf}, our neural compression approach is \textit{resolution-agnostic}. 

To summarize, the contributions of our approach are:
\begin{itemize}
    \item Firstly, we leverage neural radiance fields for extremely low-bandwidth video compression and high-fidelity video conferencing, which is resolution-agnostic. To the best of our knowledge, our approach is the first NeRF-based video compression method.
    \item Secondly, we holistically construct the framework for NeRF-based video compression and design fine-tuning embedding model to obtain fine-tuned feature as frame substitution to be transmitted for better and adaptive compression performance. 
    \item Lastly, extensive experiments demonstrate that our proposed approach can achieve resolution-agnostic and ultra-low bandwidth with high fidelity preserving for applications in video conferencing, which significantly outperforms classic video codec (HEVC) and previous model-based compression methods.
\end{itemize}

\section{Related Work}
\subsubsection{Classic Video Codec}

Many video applications utilize standard video compression modules, commonly known as codecs, including AVS, H.264/H.265 \cite{h264,hevc}, VP8 \cite{vp8}, and AV1 \cite{av1}. These codecs employ a technique that divides video frames into key frames (I-frames), capitalizing on spatial redundancies within a frame, and predicted frames (P-/B-frames), leveraging both temporal and spatial redundancies across frames. Over time, these standards have undergone enhancements, incorporating concepts like variable block sizes and low-resolution encoding \cite{av1} to optimize performance at lower bitrates.

These codecs demonstrate notable efficiency in their slow modes, if ample time and computational resources to compress videos at high quality are available. Nevertheless, for real-time applications like video conferencing, they still demand a few hundred Kbps, even at moderate resolutions such as 720p. In situations with limited bandwidth, these codecs face challenges and may only transmit at lower quality, experiencing issues like packet loss and frame corruption \cite{classic_limit}.

\subsubsection{Face Animation Synthesis}
Historically, face animation synthesis methods can be categorized into warping-based, mesh-based, and NeRF-based approaches. Among these, warping-based methods \cite{Soft-gated-warping-gan,Marionette,Liquid-warping-gan,bilayer} are particularly popular within 2D generation techniques. In these methods, source features are warped using estimated motion fields to align the driving pose and expression with the source face. For example, Monkey-Net \cite{monkeynet} constructs a 2D motion field from sparse keypoints detected by an unsupervised trained detector. Da-GAN \cite{da-gan} integrates depth estimation to enhance the 2D motion field by supplementing missing 3D geometry information. OSFV \cite{osfv} attempts to extract 3D appearance features and predict a 3D motion field for free-view synthesis.

Certain traditional approaches \cite{synthesis_obama,deferred_render} make use of 3D Morphable Models (3DMM) \cite{3dmm,3dmm_deep}, enabling a broad range of animations through disentangled shape, expression, and rigid motions. Models like StyleRig \cite{StyleRig} and PIE \cite{pie} leverage semantic information in the latent space of StyleGAN \cite{StyleGAN} to modulate expressions using 3DMM. PIRender \cite{pirender} employs 3DMM to predict flow and warp the source image. 

NeRF \cite{nerf}, a more recent method, represents implicit 3D scenes by rendering static scenes with points along different view directions, which initially gained prominence in audio-driven approaches \cite{adnerf,few_shot_talking_head,dfa-nerf} due to its compatibility with latent codes learned from audio.

\subsubsection{Neural Compression for Video Conferencing}
The limitations of classic codecs in achieving extremely low bitrates for high-resolution videos have prompted researchers to explore neural approaches for reconstructing videos from highly compact representations. Neural codecs have been specifically tailored for applications such as video streaming, live video, and video conferencing. 
%For instance, Swift is designed to compress and decompress based on residuals within a layered-encoding stack. Both NAS and LiveNAS enhance video quality using one or more Deep Neural Network (DNN) models, either at the client side for video streaming or at the ingest server for live video. These models provide adjustable knobs to control compute overheads, allowing for flexibility through the use of smaller DNNs or by adjusting the number of epochs for online fine-tuning. Across a diverse range of videos, these approaches have demonstrated improvements in bits-per-pixel consumption.

However, video conferencing presents distinct challenges compared to other video applications. Firstly, the unavailability of the video ahead of time hinders optimization for the best compression-quality trade-off.  Additionally, video conferencing content predominantly consists of facial data, allowing for a more targeted model design for generating facial videos. Several models \cite{fomm,bilayer,osfv,Maxime_low_bandwidth_video_chat,neural_compression_multi_views} have been proposed over the years, typically utilizing keypoints or facial landmarks as a compact intermediary representation of a specific pose. These representations are then used to compute the movement between two poses before generating the reconstruction. The models may incorporate 3D keypoints \cite{osfv}, off-the-shelf keypoint detectors \cite{bilayer}, or a variety of reference frames \cite{neural_compression_multi_views} to enhance prediction.

\begin{figure}[t]
    \includegraphics[width=\textwidth]{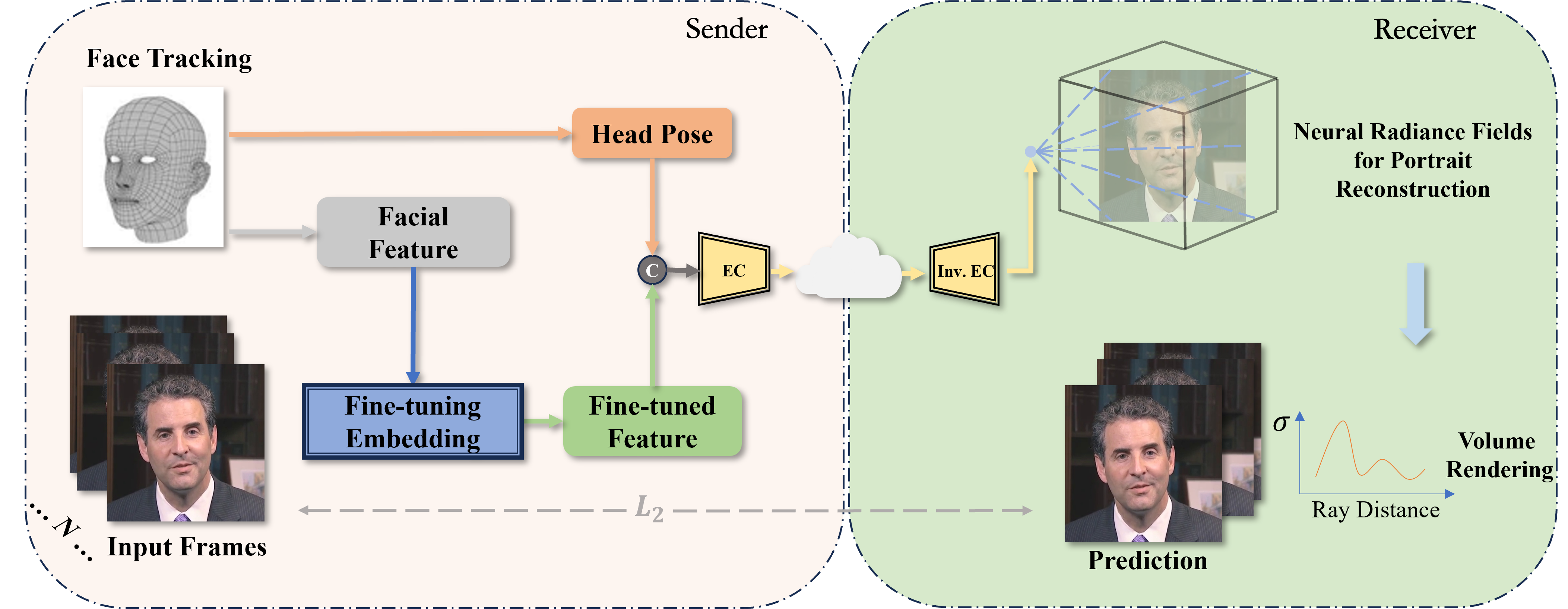}
    \caption{The overall framework of our proposed method. Face feature is extracted at the sender and substitutes frame to be transmitted with ultra-low bandwidth. At the receiver, NeRF-based model takes the received feature as input to reconstruct portrait frame. } 
    \label{overall_system}
\end{figure}

\subsubsection{Neural Radiance Field and Dynamic Rendering}

Our approach aligns with recent advancements in neural rendering and novel view synthesis, particularly drawing inspiration from Neural Radiance Fields (NeRF) \cite{nerf}. NeRF employs a Multi-Layer Perception (MLP), denoted as $F$, to acquire a volumetric representation of a scene. $F$, for each 3D point and viewing direction, predicts color and volume density. Through hierarchical volume sampling, $F$ is densely evaluated throughout the scene for a given camera pose, followed by volume rendering to generate the final image. The training process involves minimizing the error between the predicted color and the ground truth value of a pixel.

While NeRF is originally designed for static scenes, several efforts have been made to extend its applicability to dynamic objects or scenes. Some approaches \cite{adnerf,4d_facenerf,dfa-nerf} introduce a time component as input and impose temporal constraints by utilizing scene flow or a canonical frame for talking head and face animation synthesis. For example, AD-NeRF \cite{adnerf} proposes to use an audio feature as additional input with head-torso separate modeling to reconstruct natural and photo-realistic face animation. Nevertheless, NeRF-based face reconstruction model has not been proposed for video compression and video conferencing.

\begin{figure}[t]
    \includegraphics[width=\textwidth]{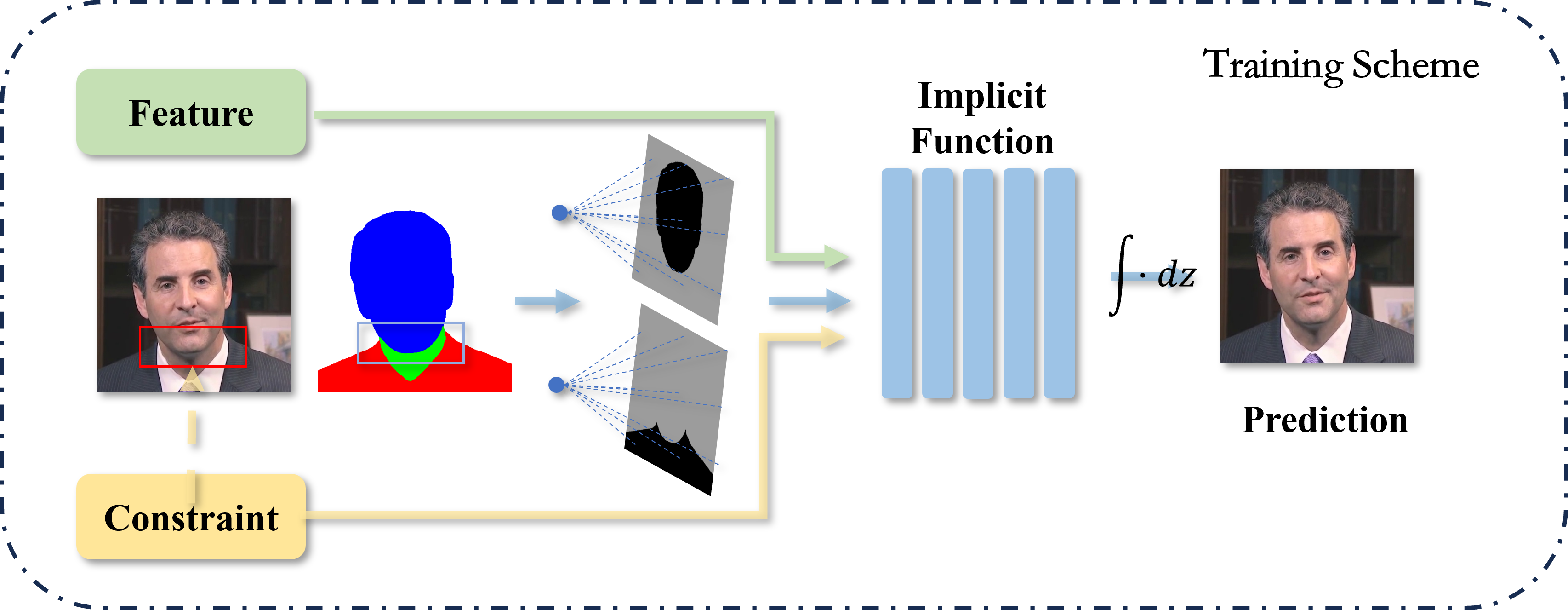}
    \caption{Training scheme of the NeRF-based reconstruction model. We leverage consistency constraint code to get better generative results.} 
    \label{training_scheme}
\end{figure}

\section{Methodology}

\subsection{NeRF-based Compression Framework}
Our objective is to leverage neural radiance fields to design a video compression framework for extremely low-bandwidth video conferencing with high fidelity. Therefore, the overall framework of our proposed approach can be regarded as a communication system which is composed of the sender, the receiver and transmission. The key insight of the proposed approach is substituting face image with feature which can be represented as low-dimensional vector for transmission. Face tracking model and entropy coding are employed for facial feature extraction and further compression at the sender before transmission. At the receiver, the face animation model based on NeRF is used to reconstruct high-fidelity and photo-realistic portrait frames from the received features. Furthermore, the overall system is end-to-end which is illustrated in Fig. \ref{overall_system}.

\subsubsection{Face Feature Extraction}
To reconstruct high-fidelity portrait frame with low-bandwidth limitation, an appropriate representation of face is essential. Rather than extracting motion keypoints in self-supervised manner described in \cite{fomm,ultra_low_fomm}, we propose to employ 3DMM \cite{3dmm,3dmm_deep} as face tracking model to extract facial expression feature and head pose for face reconstruction. 
3DMMs (3D Morphable Models) utilize a PCA (Principal Component Analysis)-based linear subspace to independently control face shape, facial expressions, and appearance. This approach allows for a disentangled representation of these facial features, enabling more flexible and intuitive manipulation of individual components. 
%By employing PCA, 3DMMs capture the variations in the data and enable the generation of a diverse range of facial shapes, expressions, and appearances through linear combinations of the learned components. 
This disentanglement is particularly valuable in applications such as face modeling and synthesis, which delivers precise control over specific aspects of the face. Therefore, we employ 3DMM as an intermediate 3D representation model to extract facial expression feature $\delta$ and head pose $p$. Following \cite{3dmm_deep}, the primitive facial expression feature can be represented as a 79-dimensional vector. In terms of head pose, a 12-dimensional vector is employed: 9 numbers for the rotation and 3 numbers for the translation.  

\subsubsection{Fine-tuning Embedding}
However, in fact, the primitive face feature extracted using the pre-trained face tracking model is still redundant. To obtain lower bandwidth transmission and better performance in compression, we leverage an attention-based encoder network \cite{attention} to construct a fine-tuning embedding of primitive feature into lower-dimensional and higher-level representation. Specifically, the fine-tuned feature used in our experiment is a 30-dimensional vector.

\subsubsection{Further Compression}
As for face feature, it's actually represented as vector with floating point values in 16 bits precision, and some classic compression schemes can be employed for further compression. Due to the characteristic of employed reconstruction model, the accuracy of the input features has a significant impact on generative performance. Therefore, the lossless compression scheme is recommended. In our approach, we compress fine-tuned face features further using \textit{Entropy Coding}. Then the coded fine-tuned face feature, together with the coded pose, are transmitted to receiver.

\subsubsection{Portrait Frame Reconstruction}
At the receiver, portrait frames are reconstructed from received face features using NeRF-based face reconstruction model, which hold the common facial expression and head poses as in \textit{source} input images. Following the recent work of Guo Y \textit{et al.} \cite{adnerf}, we employ two individual neural radiance fields to represent head part and torso part separately which demonstrates significant performance in talking-head synthesis. Nevertheless, rather than the audio feature used in \cite{adnerf}, we build the reconstruction model with facial expression feature from 3DMM as animation driving in order to maintain consistency in source and reconstructed facial expressions. Furthermore, we propose a learnable constraint to optimize the degree of fit between head and torso for better performance. More details of the reconstruction model are described in Sec. 3.2.

\subsection{Neural Radiance Fields for Face Reconstruction}
Inspired by audio driven neural radiance fields for talking-head synthesis introduced by Guo Y \textit{et al.} \cite{adnerf}, we utilize facial expression feature driven reconstruction model for neural compression. In addition to the view directions $(\theta, \phi)$ and 3D locations $(x,y,z)$, the facial expression feature $\delta$ is introduced as an additional input to the neural radiance field which is represented as an implicit function $\mathcal{N}_\Theta$. With the concatenated input vectors $(\delta, \theta, \phi, x, y, z)$, the network estimates color values $\textbf{c}$ accompanied by volume densities $\sigma$ along the dispatched rays:
\begin{equation}
    \mathcal{N}_\Theta(\delta, \theta, \phi, x, y, z) = (\textbf{c}, \sigma).
\end{equation}

\subsubsection{Consistency Constraint}
In addition, apart from the different selection of driving feature, we observe that there will be a gap between head and torso in reconstruction following the individual optimization strategy introduced in \cite{adnerf}, and thus we propose a \textit{learnable constraint code} to improve the consistency between head and torso part, which is substantiated in the ablation study of our experiments. The overall training scheme of the reconstruction model is illustrated in Fig. \ref{training_scheme}.

\subsubsection{Volumetric Rendering of Face Radiance Fields}
To generate images from this implicit geometry and appearance representation, we employ volumetric rendering. The process involves casting rays through each individual pixel of a frame, accumulating the sampled density and RGB values along the rays to calculate the final output color. Leveraging head pose tracking with 3DMM, we transform the ray sample points to the canonical space of the head model and then evaluate the dynamic neural radiance field at these locations. It's important to note that the pose $P$, obtained from head pose tracking, provides us with control over the head pose during test time. This control over head pose allows for dynamic adjustments and customization when rendering the images.

Once the color $\textbf{c}$ and volume density $\sigma$ have been predicted by the implicit function $\mathcal{N}_\Theta$, the expected color $\mathcal{C}$ of a camera ray $\textbf{r}(t) = \textbf{o} + t\textbf{d}$ with camera center $\textbf{o}$ and viewing direction $\textbf{d} = (\theta, \phi)$ is accumulated as:
\begin{equation}
    \mathcal{C}(\textbf{r};\Theta, P, \delta) = \int_{b_{near}}^{b_{far}}\sigma_\Theta(\textbf{r}(t))\cdot \textbf{c}_\Theta(\textbf{r}(t), \textbf{d})\cdot T(t)dt ,
\end{equation}
where $b_{near}$ and $b_{far}$ are near bounds and far bounds of sampling along the ray. $T(t)$ is the accumulated transmittance along the ray from $b_{near}$ to $t$:
\begin{equation}
    T(t) = exp(-\int_{b_{near}}^{t} \sigma(\textbf{r}(x))dx) .
\end{equation}
Besides, it's worth noting that we use a similar two-stage volumetric integration approach to Mildenhall \textit{et al.} \cite{nerf}.

\subsection{Optimization Details}
\subsubsection{Dataset}
We employ HDTF \cite{hdtf} as the main dataset for face animation reconstruction in the applications of video conferencing. We select videos of different identities from HDTF dataset \cite{hdtf}. There are several input resolutions for training: $128\times128$, $256\times 256$, $512\times 512$ and $1024\times 1024$. 

\subsubsection{Training Loss}
As the overall system is end-to-end, we leverage a photo-metric reconstruction error metric over the training images $I_i$ to optimize both the coarse network and fine network:
\begin{equation}
    L = \sum_{i=1}^{M} L_i(\Theta_{c}) + L_i(\Theta_f) ,
\end{equation}
where $\Theta_c$ and $\Theta_f$ are parameters of coarse and fine networks and $L_i$ is:
\begin{equation}
    L_i = \sum_{j\in pixels} \Vert \mathcal{C}(\textbf{r}_j;\Theta, P_i, \delta_i) - I_i[j] \Vert^2 .
\end{equation}
% Furthermore, as for the consistency constraint, we first select the head-torso connection region according to face parsing and then use additional pixel-wise loss for further optimizaiton:
% \begin{equation}
%     L_c = 
% \end{equation}

\begin{figure}[t]
    \includegraphics[width=\textwidth]{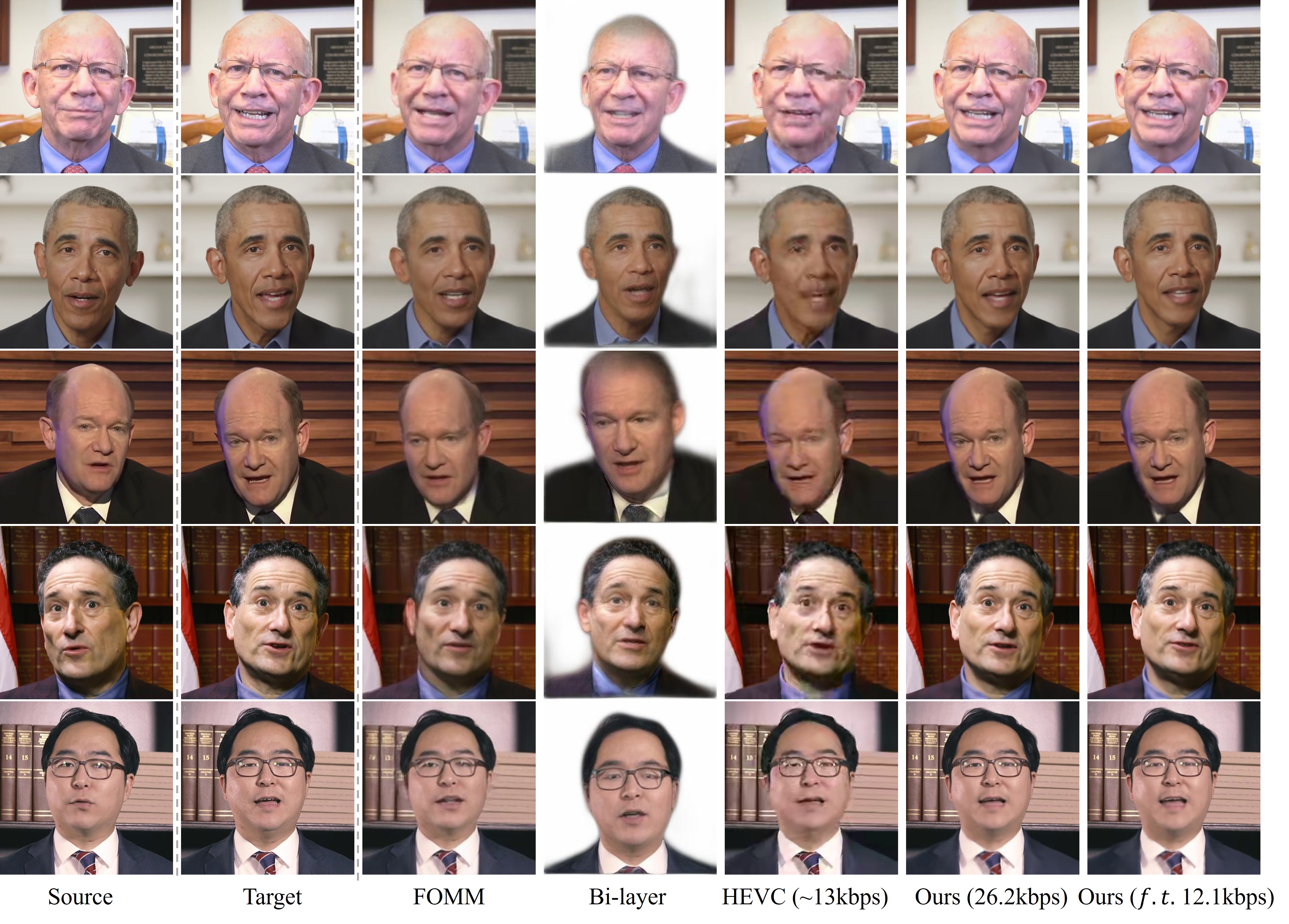}
    \caption{Qualitative results of the proposed framework compared with previous model-based compression (FOMM \cite{fomm} and Bi-layer \cite{bilayer}) and classic video codec (HEVC \cite{hevc}). Our approach, which employs NeRF-based model for high-fidelity reconstruction and feature-frame substitution for ultra-low bandwidth, outperforms other methods in image quality significantly. $f.t.$ represents fine-tuning embedding employed in the framework.} 
    \label{qualitative_eval}
\end{figure} 

\section{Experiments}
\subsection{Overview}
The goal of the proposed framework is to construct resolution-agnostic NeRF-based compression for high-fidelity portrait video conferencing with extremely low bandwidth. To demonstrate the significant performance of our approach for applications in video conferencing, we conduct both quantitative and qualitative evaluation compared with state-of-the-art model-based video compression approach and classic video codec and discuss ablation studies of our method.

\subsubsection{Metrics \& Setting}
We measure the performance of reconstruction-based models and classic codec using both quality metrics (\textbf{SSIM}, \textbf{PSNR}, \textbf{LPIPS}) and fidelity metrics. Specifically, following \cite{Marionette}, we employ \textbf{CSIM}, \textbf{AUCON} and \textbf{PRMSE} to evaluate the fidelity. Cosine similarity (\textbf{CSIM}) is used to evaluate the quality of identity preservation. \textbf{PRMSE}, the root mean square error of the head pose angles is leveraged to inspect the capability of the model to properly reenact the pose and the expression of the driver. And \textbf{AUCON} represents the ratio of identical facial action unit values between generated images and driving images. 
As for qualitative evaluation, we design the similar bandwidth of classic codec as other methods and evaluate the quality of images. In terms of quantitative evaluation, we first compare both quality metrics and bitrate-quality trade-off, which is represented as \textbf{SSIM/b.r.}, \textbf{PSNR/b.r.} and \textbf{LPIPS$\times$b.r.}, where \textbf{b.r.} represents bitrate. And then we compare the fidelity metrics and bitrate-fidelity trade-off, which is represented as \textbf{CSIM/b.r.}, \textbf{AUCON/b.r.} and \textbf{PRMSE$\times$b.r.}. Furthermore, we also demonstrate the compression performance using rate-distortion curve.

\subsection{Qualitative Evaluation}
In terms of qualitative evaluation, we compare our method with the SOTA model-based compression Bi-layer \cite{bilayer} and FOMM \cite{fomm} together with the most available and efficient classic video codec, HEVC. Specifically, we preserve the regular keypoint/landmark settings proposed in Bi-layer and FOMM and employ \textit{Entropy Coding} for further compression as well to make comparisons. For HEVC, we choose the appropriate \textit{Constant Rate Factor} to obtain similar bandwidth as our proposed method and compare the compression performance.

As illustrated in Fig. \ref{qualitative_eval}, our method can generate more realistic and high-fidelity results under extremely low bandwidth. Neither Bi-layer nor FOMM can reconstruct high-fidelity portrait due to their 2D warping based method. Classic codec HEVC has little implication on fidelity, while in similar condition (compared to ultra-low bandwidth in our method) there is much distortion that degrades the image quality. Consequently, our proposed framework delivers more appealing results for applications in video conferencing.

\begin{table}[t]
    \caption{Quantitative results over quality metrics and bitrate trade-off.}
    \centering
    \vspace{2mm}
    \begin{tabular}{cccccccc}
        \toprule
        \multirow{2}{*}{Methods} & \multicolumn{4}{c}{Quality} & \multicolumn{3}{c}{Quality-bitrate Tradeoff}\\
        \cline{2-8}
        & L1$\downarrow$~ & SSIM$\uparrow$~ & PSNR$\uparrow$~ & LPIPS$\downarrow$~  & ~SSIM/b.r.$\uparrow$~ & ~PSNR/b.r.$\uparrow$~ & ~LPIPS$\times$b.r.$\downarrow$~\\
        \midrule
        FOMM \cite{fomm}             & 0.038   & 0.77   & 24.37   & 0.12    & 0.04   & 1.41    & 2.07   \\
        Bi-layer \cite{bilayer}           & 0.23   & 0.55   & 15.88   &   0.44  & 0.014   & 0.41    & 17.23   \\
        HEVC \cite{hevc}         & 0.019   & 0.89   & 28.83   & 0.091    & \underline{0.068}   & \underline{2.22}    & 1.21   \\
        Ours   & \textbf{0.014}   & \textbf{0.95}   &  \underline{29.97}  & \textbf{0.048}    &  0.036  & 1.144    & \underline{1.19}   \\
        Ours($f.t.$)       & \underline{0.015}   & \underline{0.934}   & \textbf{30.85}   & \underline{0.05}    & \textbf{0.077}   & \textbf{2.55}    & \textbf{0.6}   \\
        \bottomrule
    \end{tabular}
    \label{quantitative_eval1}
\end{table}

\subsection{Quantitative Evaluation}
\begin{figure}[htbp]
    \includegraphics[width=\textwidth]{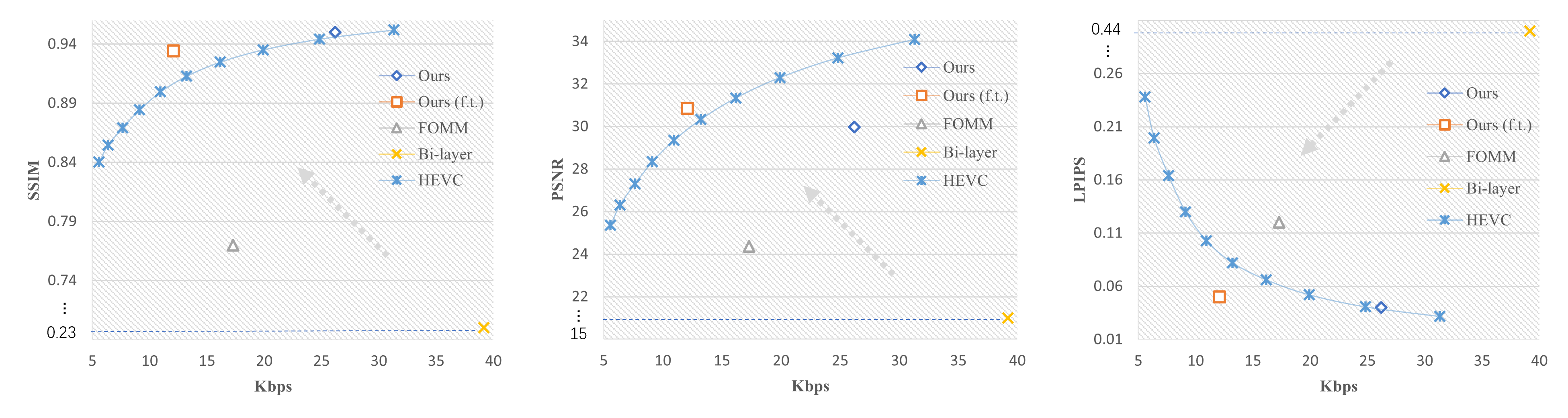}
    \caption{Rate-distortion curve for our proposed framework compared with existing model-based compression method and classic codec HEVC. The resolution for HEVC codec is $256\times 256$.} 
    \label{rate_distortion1}
\end{figure}

With regards to quantitative evaluation, we first compare both the quantitative metrics (quality and fidelity metrics) and trade-off between the bitrate and quality/fidelity represented as \textbf{SSIM/b.r.}, \textbf{PSNR/b.r.}, \textbf{LPIPS$\times$b.r.}, \textbf{CSIM/b.r.}, \textbf{AUCON/b.r.} and \textbf{PRMSE$\times$b.r.}, as shown in Table \ref{quantitative_eval1} and Table \ref{quantitative_eval2}. Furthermore, to demonstrate compression performance more clearly, we employ rate-distortion curve analysis in Fig. \ref{rate_distortion1}. 

\begin{table}[t]
    \caption{Quantitative results over fidelity metrics and bitrate trade-off.}
    \centering
    \vspace{2mm}
    \begin{tabular}{ccccccc}
        \toprule
        \multirow{2}{*}{Methods} & \multicolumn{3}{c}{Fidelity} & \multicolumn{3}{c}{Fidelity-bitrate Tradeoff}\\
        \cline{2-7}
        & CSIM$\uparrow$~ & AUCON$\uparrow$~ & PRMSE$\downarrow$~  & ~CSIM/b.r.$\uparrow$~ & ~AUCON/b.r.$\uparrow$~ & ~PRMSE$\times$b.r.$\downarrow$~\\
        \midrule
        FOMM \cite{fomm}             & 0.829   & 0.856   & 2.79    & 0.048   & 0.0495    & 48.2   \\
        Bi-layer \cite{bilayer}           & 0.518   & 0.626   &   4.86  & 0.013   & 0.016    & 190   \\
        % HEVC \cite{hevc}         & 0.932   & 0.956   & 0.091    & \underline{0.068}   & \underline{2.22}    & 1.21   \\
        Ours   & \textbf{0.956}   &  \textbf{0.989}  & \textbf{1.21}    &  0.036  & 0.0377    & \underline{31.7}   \\
        Ours($f.t.$)       & \underline{0.945}   & \underline{0.967}   & \underline{1.29}    & \textbf{0.078}   & \textbf{0.08}    & \textbf{15.609}   \\
        \bottomrule
    \end{tabular}
    \label{quantitative_eval2}
\end{table}

\subsubsection{Resolution-agnostic Analysis}
\begin{figure}[htbp]
    \includegraphics[width=\textwidth]{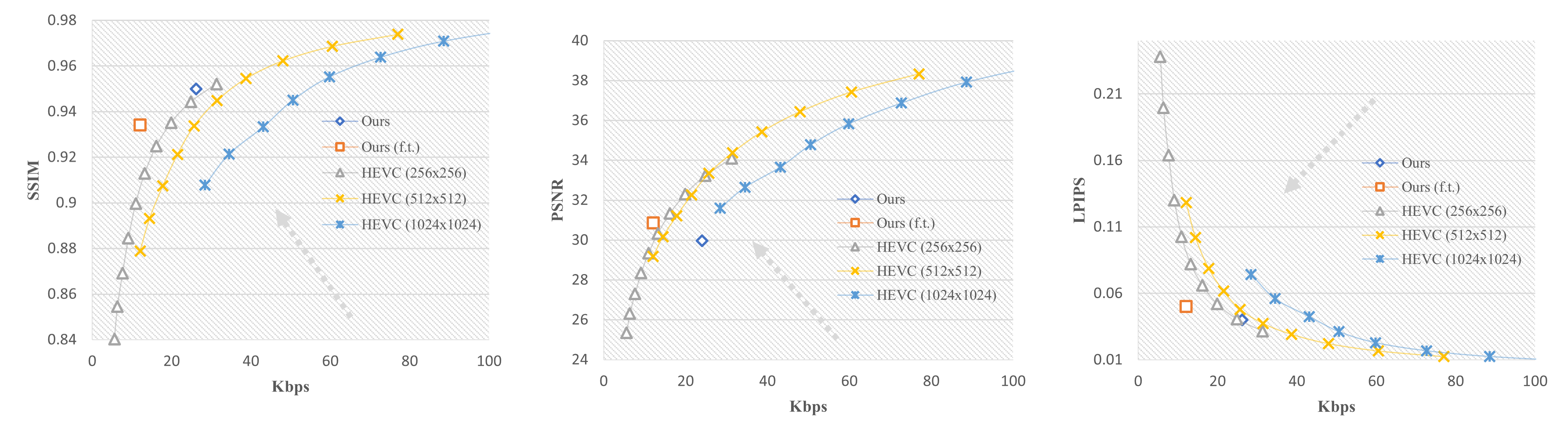}
    \caption{Rate-distortion curve of resolution-agnostic analysis for our proposed framework compared with classic codec HEVC in several different resolution settings.} 
    \label{rate_distortion2}
\end{figure}

It's worth noting that besides the significant bitrate-quality trade-off, our NeRF-based compression framework is resolution-agnostic due to the characteristic of neural radiance fields. That is the extremely low bandwidth achieved by our approach is independent of video resolution, while maintaining fidelity for higher resolution reconstruction and compression. As illustrated in Fig. \ref{rate_distortion2} rate-distortion curve, Bi-layer \cite{bilayer} and FOMM \cite{fomm} have no support for variation in resolution, while higher resolution has significant affect on performance of HEVC.

\subsubsection{Subjective Evaluation} 
\begin{figure}[htbp]
    \includegraphics[width=\textwidth]{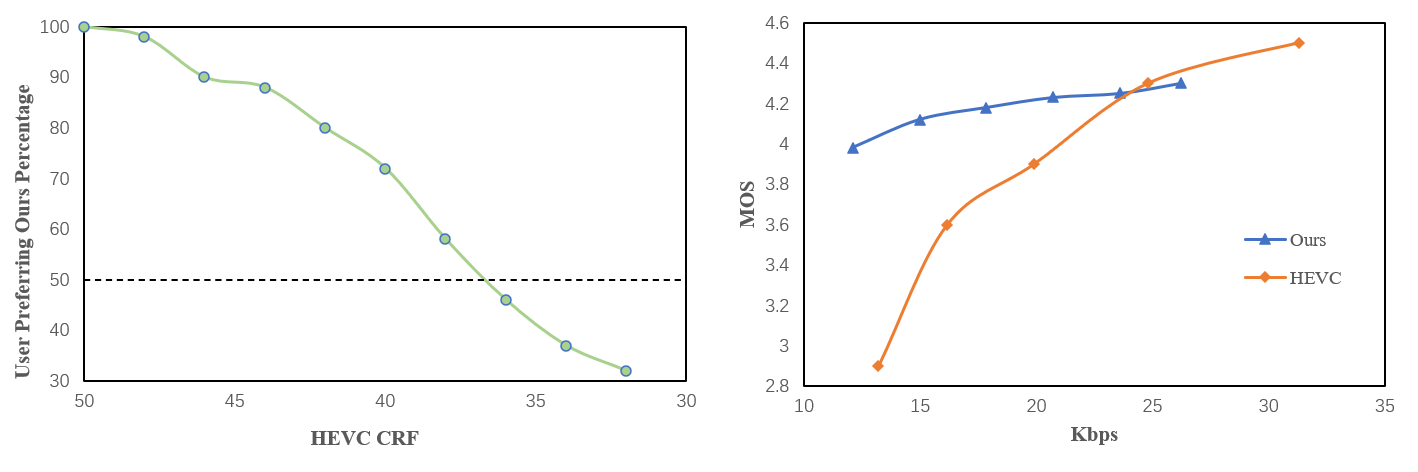}
    \caption{Subjective evaluation on the proposed framework and HEVC.} 
    \label{subjective_evaluation}
\end{figure}

Following \cite{osfv} and \cite{ultra_low_fomm}, we conduct subjective evaluation as well. We compress several clips using our framework and HEVC separately and show the compressed clips to users in video conferencing application. With various bitrate settings, we ask the users to choose the preference and compute the percentage as shown in the left side of Fig \ref{subjective_evaluation}, and to rate the clips by \textit{Mean Opinion Score} (MOS) as shown in the right side of Fig \ref{subjective_evaluation}. And in extremely low bandwidth setting, our framework shows significant performance compared to HEVC.

\subsection{Ablation Study}
We also benchmark our performance gain upon our modules. Specifically, we conduct ablations about our proposed \textit{fine-tuning embedding} model and head-torso \textit{consistency constraint} code. As for fine-tuning embedding, we have demonstrated its significant effects on video compression performance from previous experimental results, where fine-tune embedding has little affect on image quality with similar fidelity (PSNR is even higher), and reduces bandwidth significantly. The ablation study of  consistency constraint is described in Table \ref{ablation}.

\begin{table}
\caption{Ablation study on head-torso consistency constraint.}
\centering
\label{ablation}
\begin{tabular}{cccccccc}
\hline
 Setting &  L1 & SSIM & PSNR & LPIPS & CSIM & AUCON & PRMSE\\
\hline
w/o. constraint & 0.019 & 0.91 &28.68 & 0.06 & 0.94 & 0.958 & 1.31\\
w. constraint  &  \textbf{0.015}& \textbf{0.934} &\textbf{30.85}&\textbf{0.05}&\textbf{0.945}&\textbf{0.967}&\textbf{1.29}\\
\hline
\end{tabular}
\end{table}

\section{Conclusion}
In this work, we propose to leverage neural radiance fields face reconstruction model for neural video compression. Based on our NeRF-based reconstruction model, we substitute frames with features to be transmitted for video conferencing. With extensive experiments in both qualitative and quantitative aspects, we demonstrate that our novel framework implements resolution-agnostic neural compression with high-fidelity portraits in extremely low bandwidth for video conferencing, which outperforms the existing methods. As for future work, there are more further compression methods for the extracted facial feature besides lossless Entropy Coding, and we plan to leverage deep compression scheme for further feature compression to obtain better performance.

\subsubsection{\ackname} This work was supported in part by the Shanghai Pujiang Program under Grant 22PJ1406800.


\end{document}